\documentclass[11pt]{article}

\usepackage{lineno,hyperref}
\usepackage[nohyperlinks, nolist]{acronym}
\modulolinenumbers[5]
\usepackage{xcolor}
\usepackage{graphicx}
\usepackage{todonotes}
\usepackage{textcomp}
\usepackage{algorithmic}
\usepackage{url}
\usepackage{microtype}
\usepackage{amsmath,amssymb,amsfonts}
\newcommand{\footremember}[2]{
    \footnote{#2}
    \newcounter{#1}
    \setcounter{#1}{\value{footnote}}
}
\newcommand{\footrecall}[1]{
    \footnotemark[\value{#1}]
}
\usepackage[bottom]{footmisc}
\usepackage{colortbl}
\usepackage{subcaption}

\usepackage{floatpag} 

\begin{acronym}
\acro{ai}[AI]{Artificial Intelligence}
\acro{ann}[ANN]{Artificial Neural Network}
\acro{auc}[AUC]{Area under the curve}
\acro{mcc}[MCC]{Matthews correlation coefficient}
\acro{bce}[BCE]{Binary Cross-Entropy}
\acro{cnn}[CNN]{Convolutional Neural Network}
\acro{dbn}[DBN]{Deep Belief Network}
\acro{dft}[DFT]{Discrete Fourier Transform}
\acro{dl}[DL]{Deep Learning}
\acro{dnn}[DNN]{Deep Neural Network}
\acro{ecg}[ECG]{Electrocardiogram}
\acro{esn}[ESN]{Echo State Network}
\acro{fcn}[FCN]{Fully Connected Network}
\acro{fpr}[FPR]{False Positive Rate}
\acro{gaf}[GAF]{Gramian Angular Field}
\acro{gan}[GAN]{Generative Adversarial Network}
\acro{gdpr}[GDPR]{General Data Protection Regulation}
\acro{gps}[GPS]{Global Positioning System}
\acro{gru}[GRU]{Gated Recurrent Unit}
\acro{har}[HAR]{Human Activity Recognition}
\acro{iqr}[IQR]{Interquartile Range}
\acro{lstm}[LSTM]{Long Short-Term Memory}	
\acro{ml}[ML]{Machine Learning}
\acro{mlp}[MLP]{Multi-Layer Perceptron}
\acro{nlp}[NLP]{Natural Language Processing}
\acro{obs}[OBS]{OpenBikeSensor}
\acrodef{relu}[RELU]{Rectified Linear Unit}
\acro{roc}[ROC]{Receiver Operating Characteristic}
\acro{rnn}[RNN]{Recurrent Neural Network}
\acro{sas}[SAS]{Sensitivity at Specificity}
\acro{sdae}[SDAE]{Stacked Denoising Auto Encoder}
\acro{sf}[SF]{Sensor-based Fusion}
\acro{sgd}[SGD]{Stochastic Gradient Descent}
\acro{svm}[SVM]{Support Vector Machine}
\acro{tpr}[TPR]{True Positive Rate}
\acro{tsc}[TSC]{Time Series Classification}
\acro{xai}[XAI]{Explainable Artificial Intelligence}
\end{acronym}

\begin{document}

\title{CycleSense: Detecting Near Miss Incidents in Bicycle Traffic from Mobile Motion Sensors}
\author{%
  Ahmet-Serdar Karakaya\footremember{TUB}{TU Berlin, Mobile Cloud Computing Research Group, Berlin, Germany} \footremember{ECDF}{Einstein Center Digital Future, Berlin, Germany} \footnote{ask@mcc.tu-berlin.de}%
  \and Thomas Ritter\footrecall{TUB} \footnote{thomas.ritter@campus.tu-berlin.de}%
  \and Felix Biessmann \footremember{BHT}{BHT Berlin, Fachbereich VI - Informatik und Medien, Berlin, Germany} \footrecall{ECDF} \footnote{felix.biessmann@bht-berlin.de}%
  \and David Bermbach \footrecall{TUB} \footrecall{ECDF} \footnote{db@mcc.tu-berlin.de}%
  }
\date{}
\maketitle

\begin{abstract}
In cities worldwide, cars cause health and traffic problems which could be partly mitigated through an increased modal share of bicycles.
Many people, however, avoid cycling due to a lack of perceived safety.
For city planners, addressing this is hard as they lack insights into where cyclists feel safe and where they do not.
To gain such insights, we have in previous work proposed the crowdsourcing platform SimRa, which allows cyclists to record their rides and report near miss incidents via a smartphone app.

In this paper, we present CycleSense, a combination of signal processing and \acl{ml} techniques, which partially automates the detection of near miss incidents, thus making the reporting of near miss incidents easier.
Using the SimRa data set, we evaluate CycleSense by comparing it to a baseline method used by SimRa and show that it significantly improves incident detection.
\end{abstract}


\newpage
\section{Introduction\label{sec:intro}}
In recent years, more and more cities worldwide aim to reduce the modal share of car traffic in favor of bicycle traffic to reduce NO\textsubscript{x}, CO\textsubscript{2}, and particulate matter emissions, to reduce traffic jams, and to free up space that is urgently needed for other purposes ranging from vegetation that provides natural cooling in a heating world to new flats for a growing population.

A key mechanism to support this is to make bicycle traffic more attractive.
In practice, however, what keeps people from using their bikes more frequently is a lack of safety or perceived safety in cities with a car-centric traffic infrastructure~\cite{aldred2018predictors}.
Hence, city planners urgently require an overview of safety and perceived safety in their city.

Aside from actual accidents, an often overlooked aspect of cycling safety are near miss incidents\footnote{In the remainder of this paper, we will also refer to them as ``incidents''.} such as close passes or near dooring~\cite{aldred2018predictors, karakaya2020simra}.
Information on these have previously not been available or have not been easily accessible as they are distributed over private video recordings, social media posts, public CCTV footage, and other sources.
In 2019, we therefore launched the SimRa\footnote{SimRa is a German acronym for safety in bicycle traffic.} project in which cyclists record their rides and annotate them with incident information via a smartphone app~\cite{karakaya2020simra}.

While our previous approach~\cite{karakaya2020simra} already used a rudimentary heuristic for automatically detecting incidents based on acceleration sensors, it is inherently limited resulting in significant manual annotation efforts which makes it unattractive for some groups of cyclists -- in particular, elderly cyclists who have problems using smartphone apps and cyclists with significant daily mileage for whom the labeling approach is too much effort.
To increase participation, it is hence crucial to decrease manual efforts through an improved pre-detection of incidents.

As a first step towards this goal, we supervised a master's thesis which developed a neural network-based method for incident detection using the public SimRa data set\footnote{https://github.com/simra-project/dataset}~\cite{sanchez2020detecting} which, however, does not show the desired detection quality despite being an improvement over our original heuristic.
Both detection methods are currently used in the live version of the app, hence, we will later use them as baseline in our evaluation.
Besides that, there are alternative approaches for quantifying the perceived safety of bicycle traffic, e.g.,~\cite{blanc2016modeling, blanc2017safety, wu2018predicting}, but to the best of our knowledge no other methods are based on mobile motion sensory data.

Since its start in 2019, the SimRa data set has grown to more than 65,000 rides with almost 30,000 reported incidents.
These amounts of data now enable the application of much more sophisticated methods, namely \ac{ml} and/or \ac{dl}, for automatic incident detection.
Therefore, we here propose CycleSense -- a \ac{dl} model trained on smartphone sensory data from the SimRa data set to detect incidents in the SimRa app.
We hope, that it will lead to more reported incidents due to an easier reporting, and make the following contributions:

\begin{itemize}
	\item We propose an approach that combines signal processing and \ac{ml} techniques to detect incidents based on motion sensor data of cyclists with an \ac{auc} \ac{roc} of 0.906 (Section~\ref{sec:detecting}).
	\item We evaluate our approach using the SimRa data set and compare it to two baselines as well as common \acl{dl} models used in the context of \acl{tsc} (Section~\ref{sec:eval}).
	\item We discuss to which degree our approach can automate incident detection and which additional sensors are needed for full automation (Section~\ref{sec:disc}).
\end{itemize}

\section{Background\label{sec:background}}
The SimRa platform is available as a smartphone application on Android and iOS and serves as a data gathering tool for cycle tracks.

\begin{figure}[ht]
	\centering
	\includegraphics[width=0.5\textwidth]{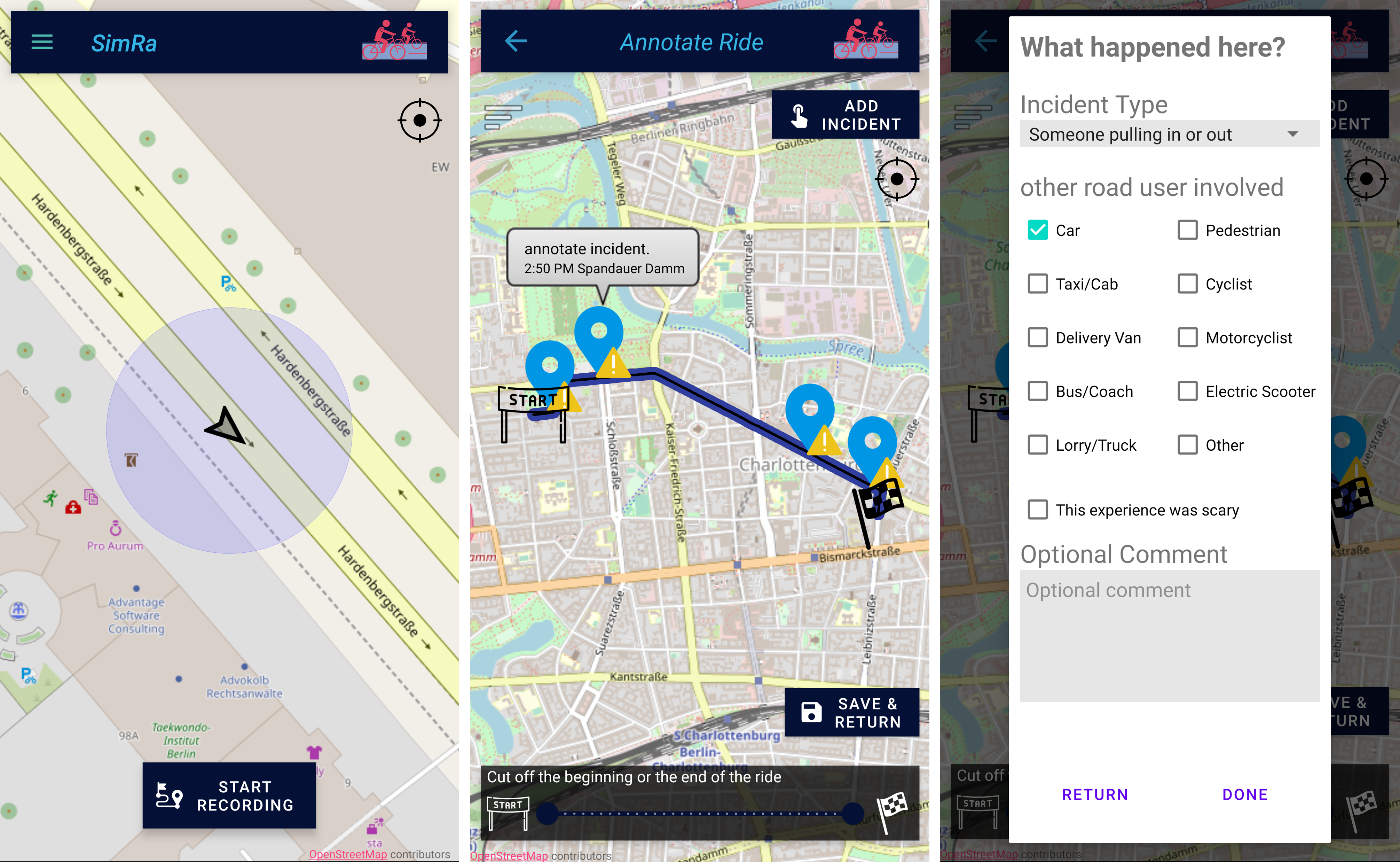}
	\caption{Screenshots of the SimRa app depicting the user story from left to right.}
	\label{fig:user-story}
\end{figure}

\textbf{User story:} In the app, users manually start recording before they begin cycling (left part of Figure~\ref{fig:user-story}).
During the ride, the \ac{gps} trace as well as accelerometer and gyroscope sensor readings are recorded.
After their ride (middle part of Figure~\ref{fig:user-story}), users can truncate their recorded route for privacy reasons and annotate the incidents represented by the blue markers.
Using the annotation menu (right part of Figure~\ref{fig:user-story}), cyclists can provide further details on incidents.
Incidents that have not been automatically detected can be added manually, false positives can be ignored.
The annotation process usually takes one to two minutes depending on the number of incidents.
Afterwards, users can upload their ride in pseudonymized form~\cite{karakaya2020simra}.

\textbf{SimRa data:} The SimRa data set consists of more than 65,000 rides from over 65 regions\footnote{https://simra-project.github.io/dashboard/}.
Berlin accounts for the majority of rides of any region: Almost half the rides and incidents have been recorded there.
Hanover and Nuremberg are the next largest regions with approximately 3500 rides each.
The SimRa data set and source code are available online and in data repositories\footnote{https://github.com/simra-project/}.
Due to per ride pseudonymization, we used only ride data for our classifier (see Section~\ref{sec:detecting}) and could not rely on profiles of individual cyclists.
Each ride file has two parts: The \textit{incident part} and the \textit{ride part}.
The \textsl{incident part} lists all incidents of the ride.
The \textsl{ride part} consists of timestamps, \ac{gps}, accelerometer, and gyroscope sensor readings.
In more recent versions of the app, linear accelerometer and rotation vector data are also included.
The recording frequencies of the various sensors (Table~\ref{tab:theoretical-measurements}) vary a lot and as a result require a partition of the SimRa data set in three distinct parts: older Android rides, newer Android rides, and iOS rides.

\begin{table}[ht]
	\centering
	\resizebox{1.0\columnwidth}{!}{%
	\begin{tabular}{cccccc}
		\hline
		& \textbf{Accelerometer} & \textbf{Gyroscope} & \textbf{\ac{gps}} & \textbf{Linear accelerometer} & \textbf{Rotation vector} \\
		\hline\hline
		\textbf{Android old} & 10 Hz & 0.33 Hz  & 0.33 Hz & \textbf{/} & \textbf{/} \\
		\hline
		\textbf{Android new} & 4 Hz & 4 Hz  & 0.33 Hz & 4 Hz & 4 Hz \\
		\hline
		\textbf{iOS} & 10 Hz & 0.33 Hz  & 0.33 Hz & \textbf{/} & \textbf{/} \\
		\hline
	\end{tabular}%
	}
	\caption{Theoretical measurement frequencies of different sensors in different parts of the SimRa data set. 
		Note that these are only the theoretical frequencies that deviate significantly from the empirical measurement frequencies that can be observed in the data set (see Section \ref{sec:disc}).}
	\label{tab:theoretical-measurements}
\end{table}

\textbf{Incident detection:} SimRa currently uses two detection methods in the live version which can be chosen by the user.
In our original approach~\cite{karakaya2020simra}, the acceleration time series is split into three second buckets.
Each bucket contains the initial GPS location and multiple sensor readings alongside their timestamp.
In each bucket, the biggest difference between two consecutive accelerometer sensor readings is calculated for each of the three directional axes.
Then, the buckets with the top two differences for each directional axis is marked as a likely incident.
Markers get placed at the GPS location of these buckets as can be seen in Figure~\ref{fig:user-story}.
The alternative approach~\cite{sanchez2020detecting} uses a simple \ac{fcn} architecture but does not consider gyroscope sensor features yet.

\section{Detecting Near Miss Incidents\label{sec:detecting}}
This section describes the process of automatically detecting incidents. 
Please note that the SimRa data are not optimized for automated processing and \ac{ml} but rather for aggregated statistics and review by humans.
Therefore, several preprocessing steps are needed (Section~\ref{subsec:preprocessing}).
We describe our \ac{ml} model in Section~\ref{subsec:architecture} and the training process in Section~\ref{subsec:training}.

\subsection{Preprocessing\label{subsec:preprocessing}}

To overcome some limitations of the SimRa data set, we use data cleaning and preprocessing steps, in a sequential multi-stage manner,  some of which are specific to some model types that will be used afterwards to classify the incidents within rides.

Before the preprocessing phase, a typical ride can be expressed by a $n \times d$ sparse matrix $X^{(i)}$, where $d$ describes the number of sensor features and $n$ represents the number of timestamps in a given ride $i \in \{1,...,R\}$.

Note that we typically only use the accelerometer, gyroscope, and \ac{gps} sensor features.
Using the linear accelerometer features in addition did not lead to a significant improvement.
Furthermore, the linear accelerometer and the rotation vector features are only available in the newer Android rides.
The non-sensory features such as phone location and bike type have a non-logical strong correlation with incidents caused by issues in the data recording phase and are therefore not used.

\begin{figure}[t]
	\centering
	\includegraphics[width=0.5\textwidth]{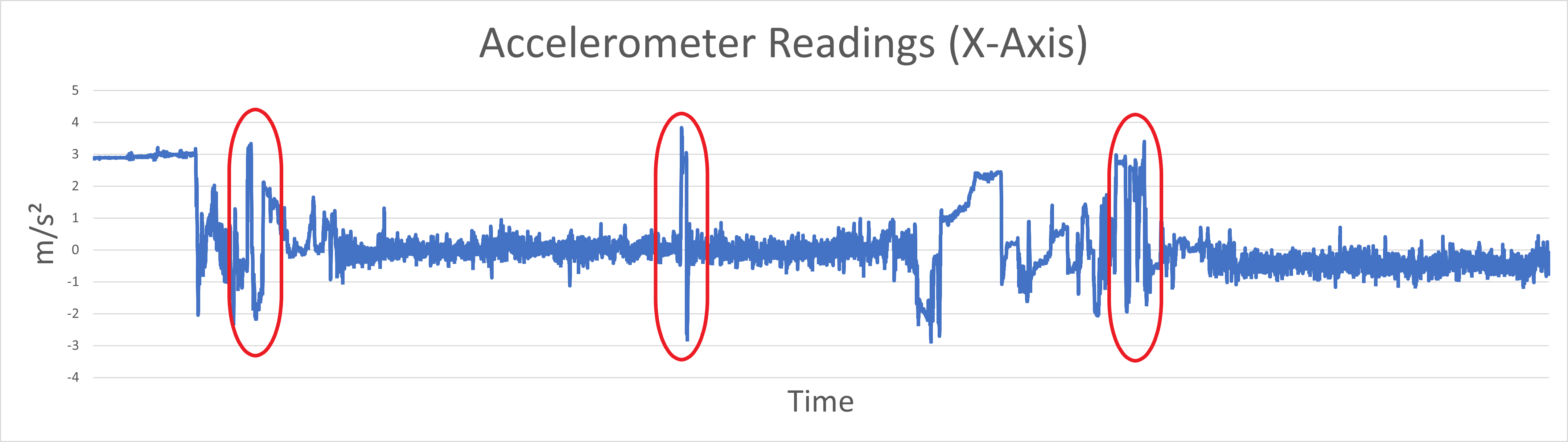}
	\caption{The accelerometer readings of an example ride. The encircled spots could indicate an incident but also driving over a curb.}
	\label{fig:x-axis}
\end{figure}

The preprocessing pipeline starts off with a manual label cleaning procedure that aims to remove some of the wrongly labeled incidents.
Identifying them solely based on time series data is practically impossible for a human.
When visualizing the accelerometer sensor readings, incidents usually stand out as sudden spikes (see Figure~\ref{fig:x-axis}), but they also could be caused by driving over a curb or suddenly stopping due to a red light.
Therefore we focus on the incidents that feature an additional description that was provided by the users.
As it is very time-consuming, this is the only manual preprocessing step, and we apply it only to the newer Android data set.
Note that this procedure did not result in a fully cleaned data set.
Next, the timestamps within rides are sorted.
Afterwards, we sort out invalid rides, i.e., rides that contain adjacent timestamps that have been recorded with a gap of more than 6 seconds.
Furthermore, we remove outliers based on the statistical definition of outliers by Tukey et al.~\cite{tukey1977exploratory} that characterizes a data point as an outlier if it fulfills one of the equations $Outlier < q_{25} - k \cdot IQR$ or $Outlier > q_{75} + k \cdot IQR$ where the \ac{iqr} is equal to the difference between the upper and lower quartiles \cite{upton1996understanding, zwillinger1999crc}.
The $k$-values we are utilizing are 1.5 for \ac{gps} outliers regarding the accuracy feature and 3.0 for velocity outliers, as we have seen reasonable results for these values.

In a further step, speed is calculated from the distance between two \ac{gps} coordinates and their respective timestamp.
Moreover, the accelerometer and gyroscope sensor data are interpolated to create equidistance over the whole time series.
This is advisable, as unevenly spaced time series data tend to pose a problem to typical \ac{ml} solutions~\cite{weerakody2021review}.
Therefore, we up-sample to a frequency of 10 Hz via linear interpolation on uniformly generated timestamps with a 100 ms interval.
That means the up-sampling factor is usually above 2.
Although some argue that interpolation is a bad solution for unevenly spaced time series data in the context of \ac{tsc}~\cite{hayashi2005covariance, eckner2012framework}, initial experiments have shown that this improves model performance.
This preprocessing stage results in dense matrices $X^{(i)}$

For better convergence of the stochastic gradient descent optimizer used in the neural network, we normalize each feature individually by its maximum absolute value.
This is nearly always an advantageous preprocessing step as it improves model stability~\cite{bishop1995neural}.

Training the model on individually labeled timestamps did not appear to be a promising approach since incidents have a certain duration, which is typically longer than 100 ms, and it is highly unlikely that the user correctly specifies the label at the exact timestamp when the incident occurred.
For that reason, we split our ride data into 10-second buckets, following a non-overlapping sliding window approach~\cite{ortiz2011dynamic}.
These buckets are then labeled in the following manner: we define a bucket as an incident bucket if any timestamp inside that bucket was labeled as an incident.
Otherwise, we define it as a non-incident bucket.

Additionally, we apply a one-dimensional $f$-point \ac{dft} on each dimension of the accelerometer and the gyroscope sensor data contained in a bucket individually.
This results in a more advanced temporal feature extraction approach that exploits the spectral power changes as time evolves by converting the time series from the time domain to the frequency domain~\cite{chen2021deep}.

To cope with the heavy label imbalance (e.g., $\approx$ 1 : 170 on rides that have recently been recorded on Android devices) that is present in the data, we use a \ac{gan} with a \ac{cnn} architecture to generate augmented data and thereby lower the imbalance gap by 10\% as this has shown to produce good results in our experiments.
The aforementioned $f$-point \ac{dft} is applied on these synthetic incident buckets as well.

\subsection{Model Architecture\label{subsec:architecture}}
As our problem setting is similar to the \ac{har} task (see Section~\ref{sec:rw}), we build a customized \ac{ann} inspired by the DeepSense architecture proposed by Yao et al.~\cite{yao2017deepsense}.

In a first step, the network input is split based on the sensor that has produced it into accelerometer, gyroscope and \ac{gps} (i.e., velocity).
Simultaneously, the previously Fourier transformed accelerometer and gyroscope data are separated into their real and imaginary parts.

Then, \ac{sf} is applied, a method that considers each sensor individually in order to extract sensor-specific information~\cite{elmenreich2002sensor}. 
Furthermore, it also enables the application of different individual subnets that are varying in complexity for each sensor input.
Each subnet has three convolutional layers that use 64 kernels, kernel sizes between $(3,3,1)$ and $(3,3,3)$, and a stride size of $1$.
While in the first convolutional layer no padding is used, the second and third convolutional layers apply zero-padding which differs from the original DeepSense framework proposed by Yao et al.~\cite{yao2017deepsense}. 
Another difference is that we use 3D-convolution instead of the 2D- and 1D-convolution that were applied in the original model. 
3D-\acp{cnn} are more suitable for detecting spatiotemporal features compared to 2D-\acp{cnn}~\cite{tran2015learning}. 
The described convolutional layers are complemented by batch normalization layers to reduce internal covariate shifts~\cite{ioffe2015batch}, by \ac{relu} activation, and by Dropout layers for regularization.

Our addition of residual blocks is also a slight modification of the original framework. 
The reasoning behind that change is that, in some cases, deeper models might have difficulties in approximating identity mappings by multiple nonlinear layers~\cite{he2016deep}. 
Residual blocks have been applied with great success to overcome this issue~\cite{he2016deep}.

\begin{figure}[t]
\centering
    \includegraphics[width=0.4\columnwidth]{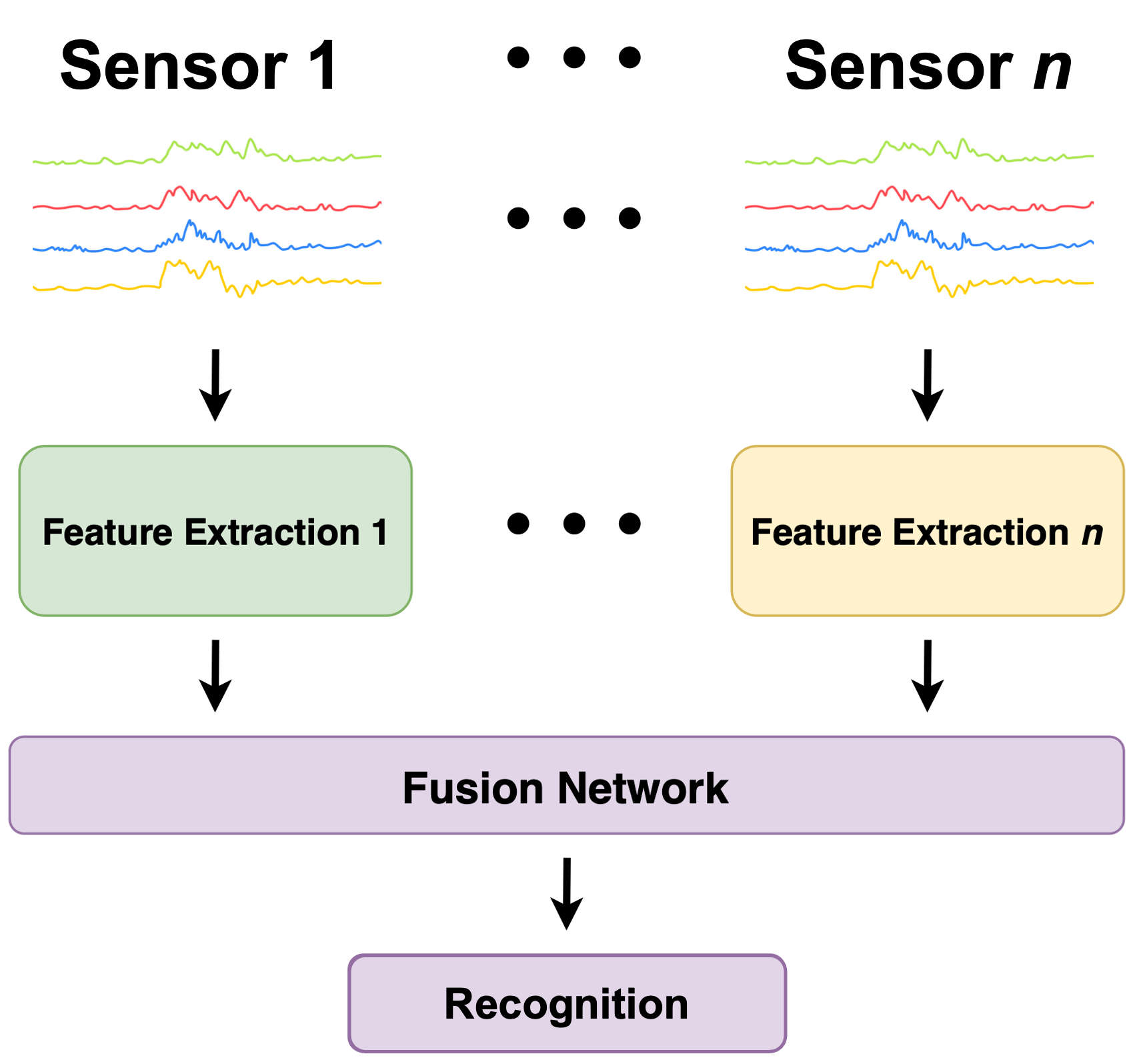}
    \caption{The model architecture in a nutshell: Using subnetworks for feature extraction of the different sensors (accelerometer, gyroscope, and \ac{gps}) and afterwards a fusion network to combine the features for final incident recognition~\cite{chen2021deep}.}
    \label{fig:sfn}
\end{figure}

Next, the outputs of the different subnets are merged in a convolutional fusion network. 
Its architecture is similar to the individual subnets containing six convolutional layers, residual blocks, batch normalization, \ac{relu} activation, and Dropout. 
The full process is shown in Figure~\ref{fig:sfn}.

The last big component of CycleSense is a \ac{rnn}.
\ac{rnn} architectures such as \ac{lstm}~\cite{hochreiter1997long} or \acp{gru}~\cite{chung2014empirical} are capable of holding information the network has seen before and using it to make predictions in the current state. 
In doing so, it is possible to identify patterns or relationships inside the timestamps of a bucket or between buckets. 
Similar to Yao et al.\ \cite{yao2017deepsense}, we also chose stacked \ac{gru} cells as they efficiently improve the model capacity~\cite{goodfellow2016deep}.

\begin{figure}[t]
	\centering
	\includegraphics[width=\textwidth]{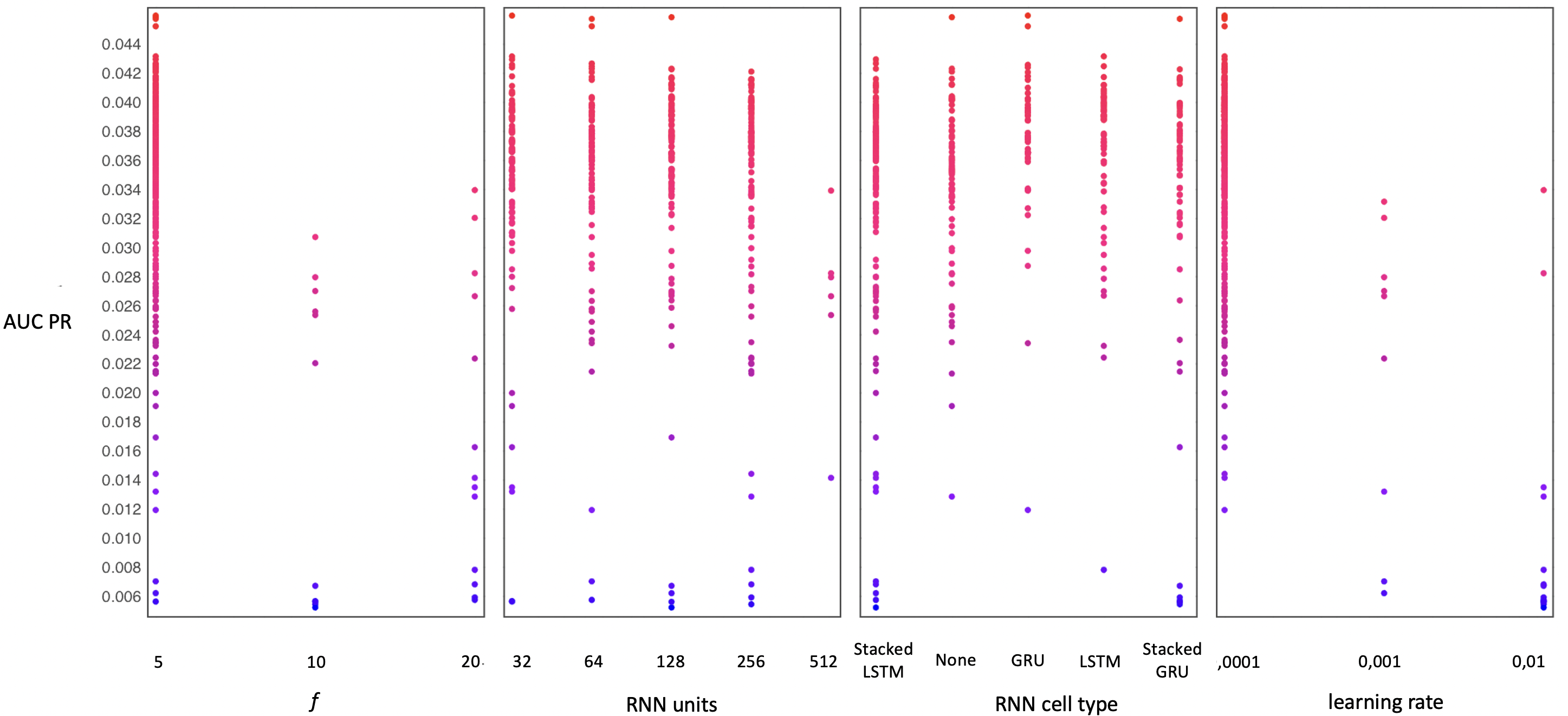}
	\caption{Results of the hyperparameter optimization for the four variables $f$ (the window size), the number of \ac{rnn} units used, the \ac{rnn} cell type utilized, and the learning rate from left to right.}
	\label{fig:hpo}
\end{figure}

To determine the optimal set of parameters for training CycleSense, we have conducted a grid search on a variety of hyperparameters, some of which are shown in Figure~\ref{fig:hpo}.

In the following, we use \ac{gps}, accelerometer, and gyroscope data as model input if not indicated otherwise.  
Linear accelerometer data was only used in a few experiments as it is not available in our iOS and older Android data sets.
Our implementation of CycleSense is available on GitHub\footnote{https://github.com/simra-project/CycleSense}.

\subsection{Model Training\label{subsec:training}}

For model training, the data set was split randomly into a training set (60\%), a validation set (20\%), and a test set (20\%).
Furthermore, the exact same splits are used for each model to improve comparability.
We trained the final model for 60 epochs on an NVIDIA K80 GPU.
We utilized a \ac{bce} loss function that was updated with Adam optimization, a \ac{sgd} method, a learning rate of 0.0001, and early stopping with a patience value of 10 epochs on the \ac{auc} \ac{roc} of the validation set.
It is important to note that we did not use the complete SimRa data set.
Instead, we used only a smaller subset of more recent rides recorded on Android devices since the heterogeneity of the data set across different versions and operating systems (see Section \ref{sec:disc}) did not allow us to train a model properly on the full data set.
In addition, due to limited access to hardware, we focused predominantly on data originating from the Berlin region if not stated otherwise.

As previously described, one notable challenge we were facing was the extreme label imbalance present in the data.
This is due to the fact that incident buckets are far rarer than non-incident buckets.
To cope with that, we trained our model by using a weighted loss function with the class weights of the train data set as weights.
For example, the class weights were 1 and 170 for the rides that have recently been recorded on Android devices.

While the DeepSense model was trained in a standard fashion, we use stacking during the training of CycleSense.
Stacking (or stacked generalization) is an ensemble learning method that combines the predictions of several different models in order to contribute equally to a collective prediction.

However, we are also not interested in equal contributions of the network since that could overvalue models with a poor performance.
We therefore changed the CycleSense model to an integrated stacking model by adopting the idea of stacked generalization~\cite{wolpert1992stacked}, where the fusion network acts as the meta-learner.
Also, we deviate from a pure stacking model.
This is the case, as the meta-learner does not get any classification output of the subnetworks as input aside from the latent features in the last layer of the subnetworks.
Thereby, the weights of the submodel layers that have been pretrained individually are loaded and frozen, so they are not updated during the training of the whole CycleSense model.
This learning procedure further improved our results as shown in Section~\ref{sec:disc}.

\section{Evaluation\label{sec:eval}}
To evaluate CycleSense' training results, we have to put them into context. 
For this purpose, we compare them to the two detection methods currently used in the app as discussed in Section~\ref{sec:background}.
We give an overview of the changes we made to the baseline methods with the goal of a fair comparison in Section~\ref{subsec:baselines}.
We also describe the metrics that we use to compare our model to the baseline methods (Section~\ref{subsec:metrics}) before presenting the results of our evaluation (Section~\ref{subsec:eval-results}).

\subsection{Baselines\label{subsec:baselines}}

The first baseline is our original heuristic~\cite{karakaya2020simra} which is based on the underlying assumption that incidents will often result in sudden acceleration spikes, e.g., when braking or swerving to avoid obstacles.
We made some small changes to this heuristic to enable its compatibility with the \ac{auc} \ac{roc} metric, thus, increasing the comparability with our approach.

As a second baseline, we retrained the \ac{fcn} model from the alternative approach~\cite{sanchez2020detecting}.
We used the original preprocessing pipeline (which differs significantly from the here presented one) but used the full data set as introduced in Section~\ref{sec:background}.
We skipped the under-sampling step, disregarded the phone location and the bike type feature for the reasons mentioned in Section~\ref{subsec:preprocessing}, and used a non-overlapping sliding window approach with 10 second windows for better comparability.

The third baseline is DeepSense~\cite{yao2017deepsense}, which we implemented and trained as the authors describe in their work.
For the differences between DeepSense and CycleSense, see sections~\ref{subsec:architecture} and~\ref{subsec:tsc}.

Based on these changes for improved comparability, we retrained the original model.
We use both baselines for comparison as they are, to our knowledge, the only approaches for (semi-)automatically detecting incidents based on sensory time series data.
Furthermore, they have been developed on the SimRa data set, which enables a fair comparison.

\subsection{Metrics\label{subsec:metrics}}

Due to the massive label imbalance already mentioned earlier, common metrics such as accuracy, F1-score, and precision are difficult to interpret.
Moreover, in our scenario it is more important to find the true near miss incidents than to classify non-incidents correctly, as False Positives can be more easily corrected by the user of the SimRa app.
For both reasons, a high number of False Positives is more acceptable than a low number of True Positives, which further limits the usefulness of such metrics like precision, F1-score, or \ac{mcc}.
Therefore, we focus on the \ac{auc} of the \ac{roc} metric, which is insensitive to changes in class distribution~\cite{fawcett2006introduction} while also reporting the respective confusion matrices.

\subsection{Evaluation Results\label{subsec:eval-results}}

\begin{figure}[t]
	\centering
	\includegraphics[width=0.8\textwidth]{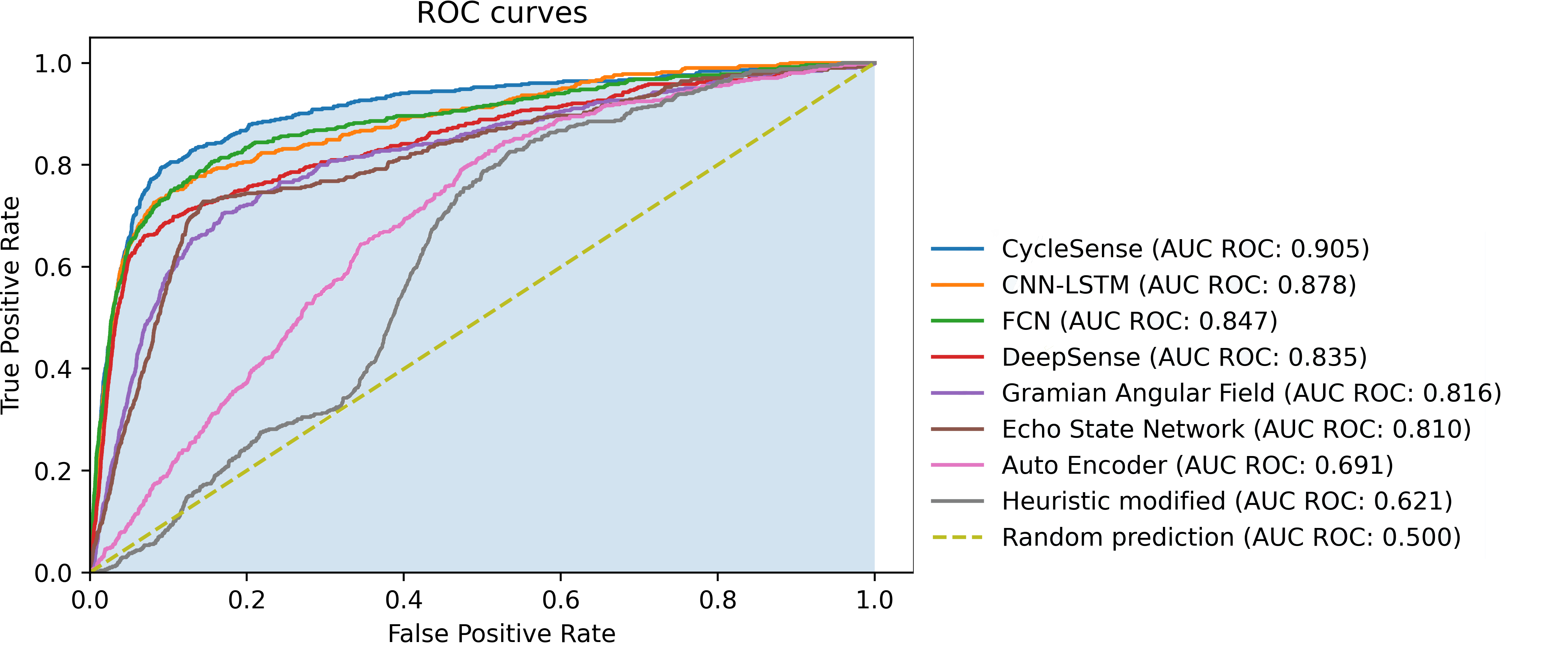}
	\caption{Comparison of the baselines and common model architectures used in the context of \ac{dl} for \ac{tsc}~\cite{ismail2019deep} with the CycleSense model. Note that all of these models have been trained and evaluated on rides contained in the SimRa data set that have been recorded with more recent versions of the SimRa Android app.}
	\label{fig:roc-auc-results}
\end{figure}

In a first step, we compare CycleSense to the two baselines and common model architectures used in the context of \ac{dl} for \ac{tsc}~\cite{ismail2019deep}.
All of these were trained on the Android data set consisting of more recent rides which provides the best results for all approaches.
Figure~\ref{fig:roc-auc-results} and Table~\ref{tab:roc-auc-results} show the differences in performance.

The \ac{fcn} and CycleSense clearly outperform the modified heuristic (0.621 \ac{auc} \ac{roc}).
However, there is still a big performance gap between our model and the \ac{fcn} model.
While the \ac{fcn} model scores 0.847 \ac{auc} \ac{roc}, CycleSense achieves an \ac{auc} \ac{roc} score of 0.906, i.e., there is a chance of $\approx$ 90.6\% that the model can distinguish correctly between a randomly chosen incident and non-incident bucket.
Furthermore, our model performs better than other model architectures that are common for \ac{dl} in \ac{tsc}~\cite{ismail2019deep}: Auto Encoder, \ac{gaf}, \ac{esn}, and the \ac{cnn}-\ac{lstm} model.
With regard to the increasing model complexity, we clearly see diminishing returns.
We can see this in the example of the rather simple \ac{cnn}-\ac{lstm} model which exhibits a relatively close performance to the much more complex CycleSense model with stacking.
The \ac{cnn}-\ac{lstm} model has $\approx$ 90,000 parameters, while the CycleSense model has $\approx$ 1,100,000 parameters.
As a consequence, the time to evaluate the test set of the newer Android data consisting of 795 rides took 4 seconds with the \ac{cnn}-\ac{lstm} model and 56 seconds using CycleSense on the NVIDIA GPU.

\begin{table}
	\centering
	\resizebox{\columnwidth}{!}{%
	\begin{tabular}{cccccccccc}
		\hline
		\centering
		& \textbf{TN} & \textbf{FP} & \textbf{FN} & \textbf{TP} & \textbf{\ac{auc} \ac{roc}} & \textbf{Precision} & \textbf{Recall} & \textbf{F1-Score} & \textbf{MCC} \\
		\hline\hline
		\textbf{CycleSense} & 107934 & 10893 & 104 & 400 & \cellcolor{yellow}0.906 & \cellcolor{yellow}0.035 & 0.794 & \cellcolor{yellow}0.068 & \cellcolor{yellow}0.156 \\
		\hline
		\textbf{CNN-LSTM} & 106427 & 12400 & 127 & 377 & 0.878 & 0.030 & 0.748 & 0.057 & 0.135 \\
		\hline
		\textbf{FCN} & 100932 & 18500 & 98 & 402 & 0.847 & 0.021 & \cellcolor{yellow}0.804 & 0.041 & 0.115 \\
		\hline
		\textbf{DeepSense} & 106192 & 12635 & 153 & 351 & 0.835 & 0.027 & 0.696 & 0.052 &  0.123 \\
		\hline
		\textbf{GAF} & 98782 & 20045 & 150 & 354 & 0.816 & 0.017 & 0.702 & 0.034 & 0.092 \\
		\hline
		\textbf{ESN} & 101670 & 17157 & 138 & 366 & 0.810 & 0.021 & 0.726 & 0.041 & 0.107 \\
		\hline
		\textbf{Auto Encoder} & 58416 & 60411 & 88 & 416 & 0.691 & 0.007 & 0.825 & 0.014 & 0.041 \\
		\hline
	\end{tabular}%
}
\caption{Comparison of the baselines and common model architectures used in the context of \ac{dl} for \ac{tsc}~\cite{ismail2019deep} with the CycleSense model. See the discussion in Section~\ref{subsec:metrics} about the usefulness of various metrics. Note also that all of these models have been trained and evaluated on rides contained in the SimRa data set that have been recorded with more recent versions of the SimRa Android app. For all models the threshold which optimizes Youden's index\cite{youden1950index} were chosen.}
\label{tab:roc-auc-results}
\end{table}

In another experiment, we include the linear accelerometer sensor values in addition to the accelerometer, gyroscope and \ac{gps} data we used so far.
The result for CycleSense is again an \ac{auc} \ac{roc} of 0.906, although the model requires more memory, training and processing time.
Therefore, we leave out the linear accelerometer feature.

\begin{figure}[t]
	\centering
	\begin{subfigure}[b]{0.475\textwidth}
		\centering
		\includegraphics[width=\textwidth]{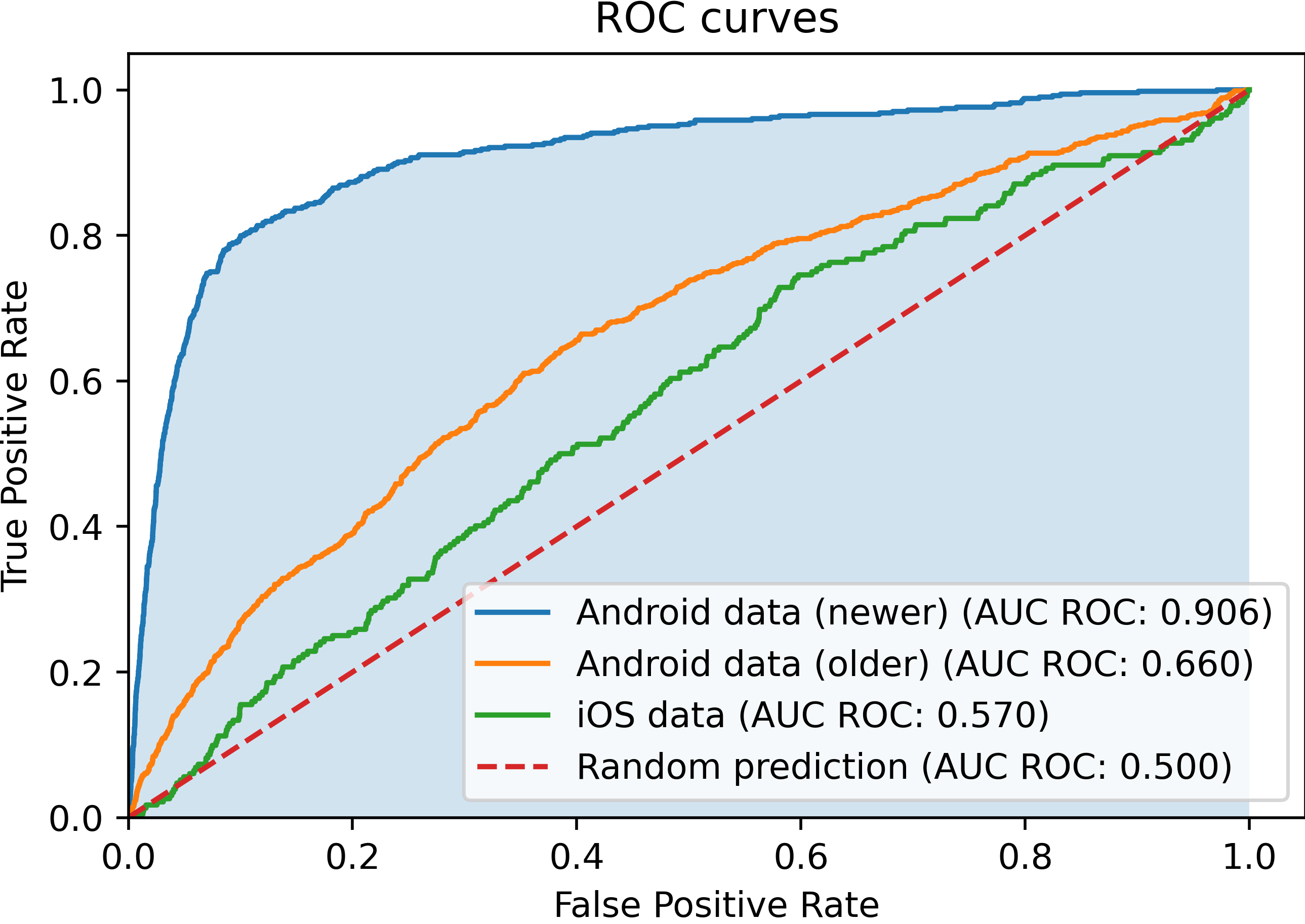}
		\caption{\small CycleSense trained only on the newer Android rides and evaluated on all parts of the data set.}
		\label{fig:traindonone}
	\end{subfigure}
	\hfill
	\begin{subfigure}[b]{0.475\textwidth}
		\centering
		\includegraphics[width=\textwidth]{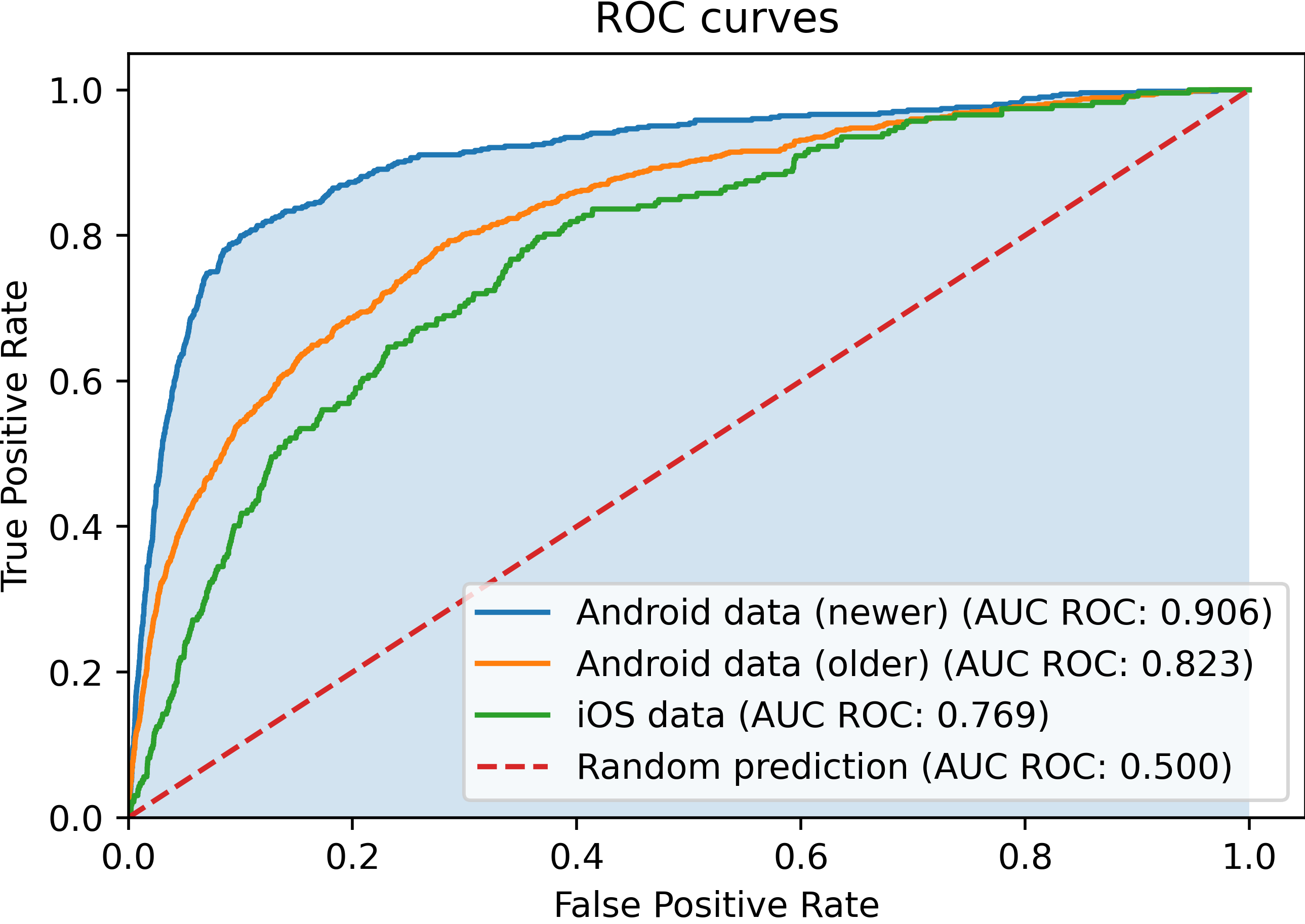}
		\caption{\small CycleSense trained and evaluated individually on all parts of the data set. \newline}
		\label{fig:individually}
	\end{subfigure}
	\caption{Comparison of CycleSense trained only on the newer Android rides or individual CycleSense models trained for each part of the data set. Note that only the newer Android data set was manually cleaned.}
	\label{fig:comp-trainedonone-individually}
\end{figure}

So far, we have predominantly focused on newer rides recorded on the Android version of the SimRa app.
As shown in Figure~\ref{fig:traindonone}, the model performs far worse on the other splits of the data set.
Therefore, it was necessary to train an individual model for each part of the data set. 
This yields far better results (Figure~\ref{fig:individually}).

As the Berlin region has by far recorded the most rides, we have so far used only those for tuning, training, and evaluating our model.
The SimRa app, however, is deployed in many more regions, so our model is clearly required to perform there, too.
For this reason, we have evaluated the Berlin CycleSense model on the newer Android rides coming from Hanover and Nuremberg.

\begin{figure}[t]
	\centering
	\begin{subfigure}[b]{0.475\textwidth}
		\centering
		\includegraphics[width=\textwidth]{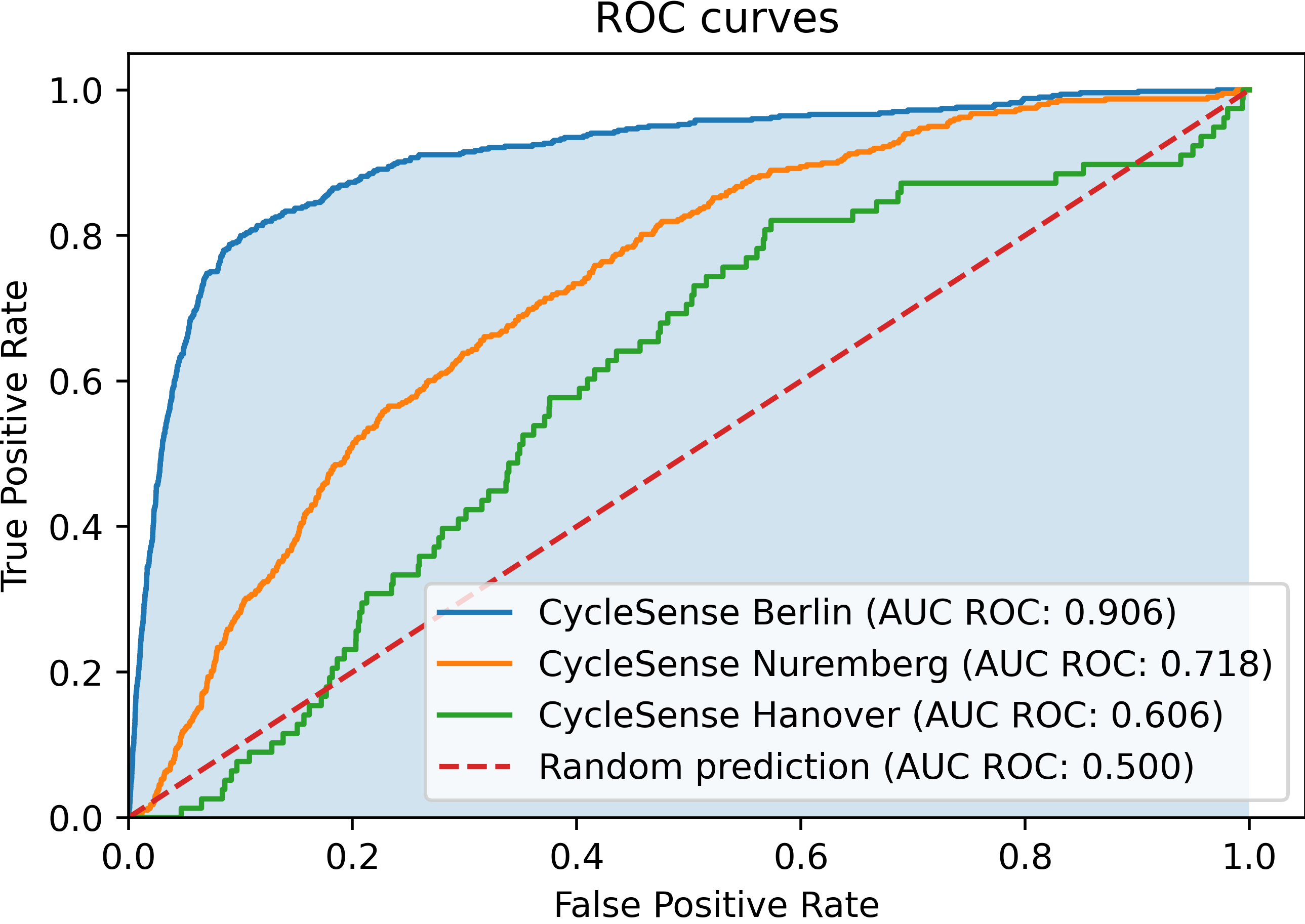}
		\caption{\small Performance of CycleSense trained on the new Android data set of rides recorded in Berlin. \newline}
		\label{fig:different-city-trained-on-berlin}
	\end{subfigure}
	\hfill
	\begin{subfigure}[b]{0.475\textwidth}
		\centering
		\includegraphics[width=\textwidth]{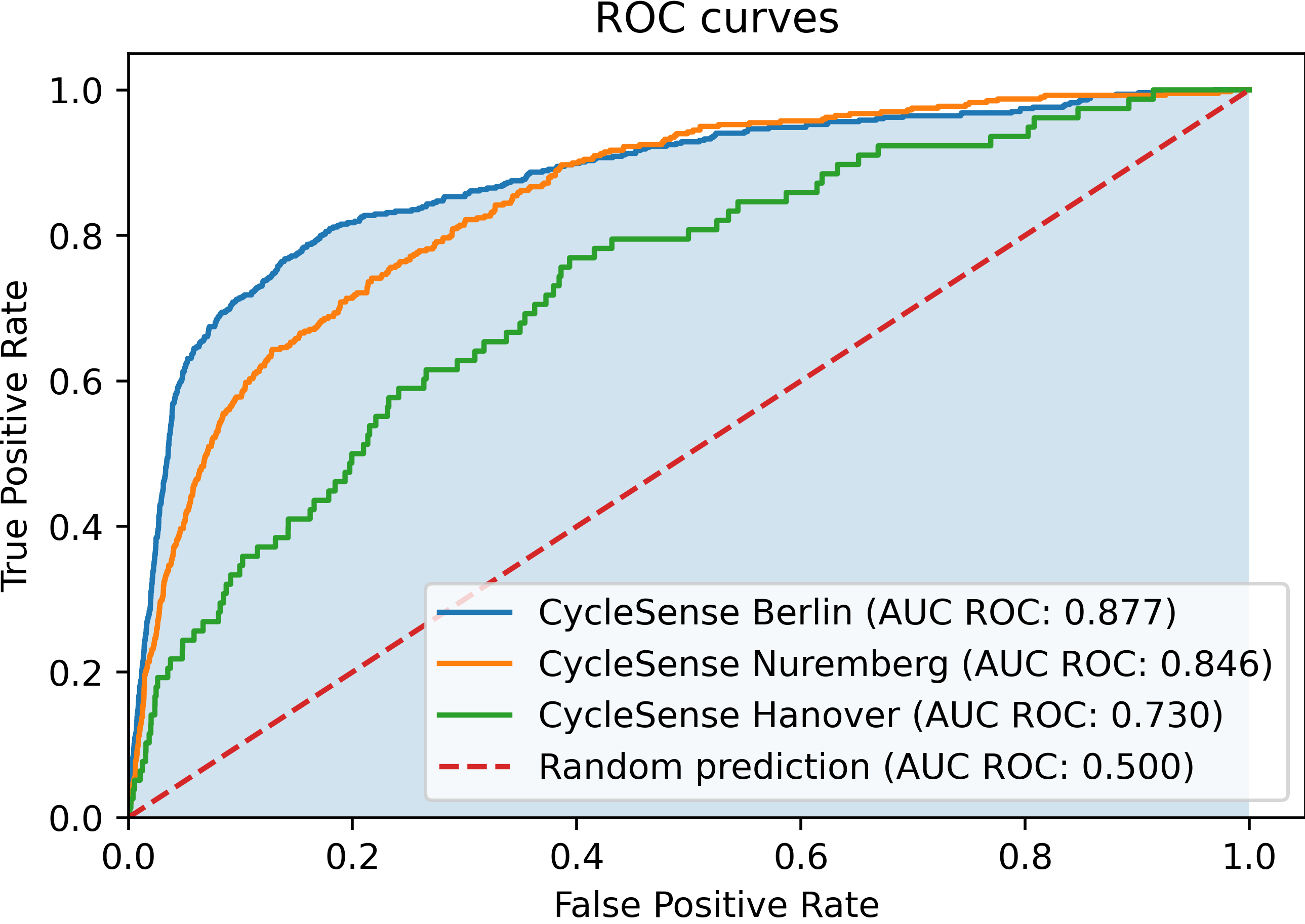}
		\caption{\small Performance of CycleSense trained on the new Android data set of rides recorded in Berlin, Nuremberg, and Hanover.}
		\label{fig:different-city-trained-individually}
	\end{subfigure}
	\caption{Comparison of the performance of CycleSense trained on Berlin and trained on the other regions combined.}
\end{figure}

The outcome of this experiment is visualized in Figure~\ref{fig:different-city-trained-on-berlin}.
It clearly shows, that CycleSense does not perform as good as in Berlin.
Since most regions lack training data, it would not be a feasible solution to train models individually per region in the current stage of the SimRa project.
Instead, we retrain CycleSense on a data set that includes all the rides recorded on the newer versions of the Android app within the Berlin, Nuremberg, and Hanover regions.
The results from Figure~\ref{fig:different-city-trained-individually} show that this clearly improves the performance in these additional regions.
At the same time, the performance on the Berlin data set has declined only slightly (0.029 \ac{auc} \ac{roc}) by comparison.
Nevertheless, the \ac{auc} \ac{roc} for Berlin is still the highest and clearly above Nuremberg, which is well ahead of Hanover.

\section{Discussion\label{sec:disc}}
Our results demonstrate that the proposed CycleSense model outperforms every other model that we have compared to.
While improving the model's architecture and other components, we identified a number of challenges inherent to the problem setting that we discuss in the following.

\subsection{Impact of Preprocessing \& Training Steps}

\begin{figure}[t]
	\centering
	\includegraphics[width=0.5\textwidth]{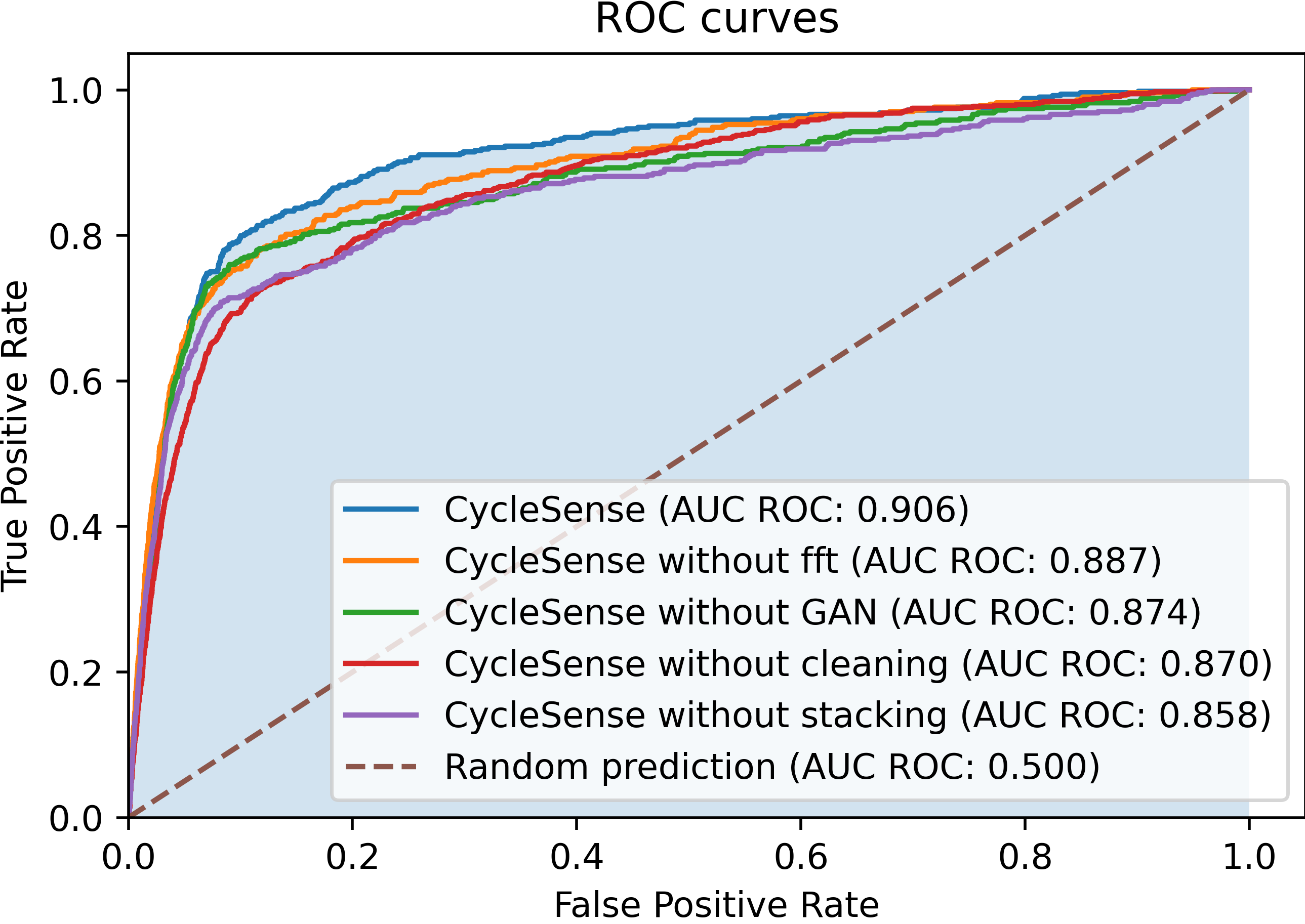}
	\caption{Impact of different preprocessing and training steps on the performance of CycleSense.}
	\label{fig:impact}
\end{figure}

In a first step, we want to highlight the importance of our preprocessing and training methods for the success of our model.
Therefore, Figure~\ref{fig:impact} illustrates the performance of CycleSense when one of the preprocessing or training methods is skipped, resulting in a significant drop in performance in each of the four examples.

\subsection{Limitations of Crowdsourced Data}

Crowdsourced data are generally noisy which could contribute to a reduced model performance.
They do also contain several biases, e.g., the Selection Bias, Device Positioning Bias, Extreme Aversion Bias, or Confirmation Bias~\cite{basiri2019crowdsourced, chakraborty2017makes, kahneman1991anomalies}.
In this context, the heterogeneity across devices and platforms is likely another factor of influence.
As already described, the application that is used by contributors to record their rides is available on two platforms. 
Those platforms are supported by a wide range of different devices and models with different hardware inside. 
Phone manufacturers use different \ac{gps}, gyroscope, and accelerometer sensors that can vary immensely in sensitivity and overall behavior.
Stisen et al.~\cite{stisen2015smart} have shown that there is major heterogeneity when it comes to the accuracy of sensor readings. 
While the main focus of that study was on accelerometer data, Kuhlmann et al.~\cite{kuhlmann2021smartphone} have shown that orientation sensor data is also affected by this variability.
This is in contrast to the \ac{har} tasks, we compare our task to.
The \ac{har} data set~\cite{anguita2013public} was generated under laboratory conditions that always used the same smartphone type body mounted to the exact same position on selected study participants.
This significantly simplifies the classification task.
Another factor that contributes to the issue of noisy crowdsourcing data in the SimRa data set is the fact that some users misinterpret the purpose of the SimRa app.
Instead of labeling incidents, they report, e.g., dangerous areas or annoying traffic lights.
These can of course not be captured by the sensors used.
Many such ``incidents'' can be identified through the comment column of the data set.
It should also be noted that the SimRa data set was not recorded and labeled with the goal of developing ML models -- the goal was to record data which will be analyzed and processed by humans.

\subsection{Technical Limitations of the SimRa Data Set}
\label{subsec:technical_limitations}

\begin{figure}[t]
	\centering
	\includegraphics[width=0.45\textwidth]{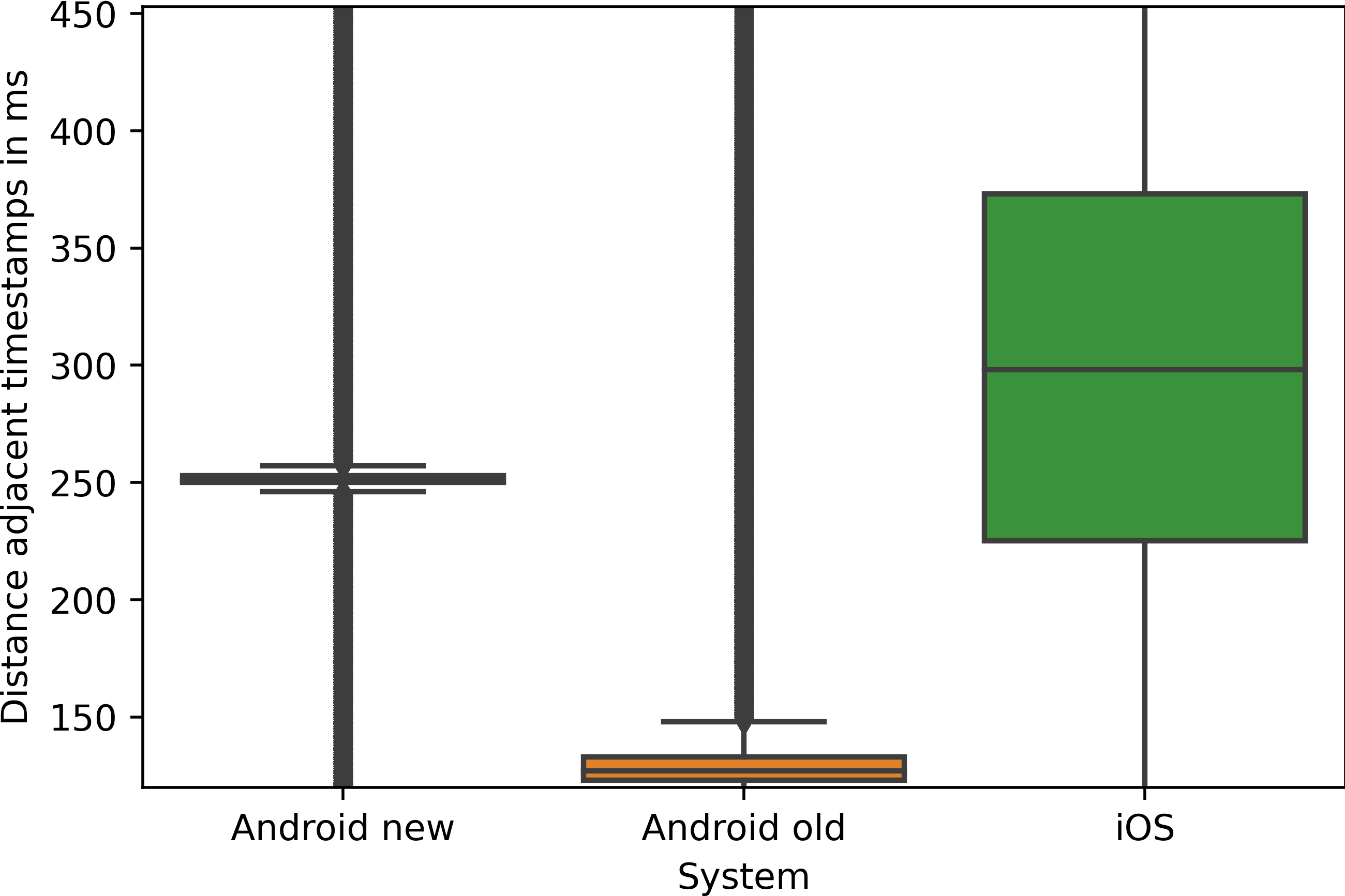}
	\caption{Box plot showing the empirical measurement times observed in the SimRa data set for rides recorded with different versions of the SimRa app.}
	\label{fig:emp-measurements}
\end{figure}

Recording rates of sensor readings deviate among devices and operating systems and impair model predictions~\cite{stisen2015smart}, this may also affect the performance of CycleSense:
As illustrated in Figure~\ref{fig:emp-measurements}, the recording rates differ significantly in older and newer Android rides as well as in iOS rides.
While the iOS rides' median ($\approx$ 300 ms) is similar to the one of newer Android rides, the \ac{iqr} is much greater and spans approximately 150 ms.
This circumstance could indicate that the relatively weak model performance on this data set can partly be explained by that factor. 
Furthermore, the gyroscope data is recorded with a higher frequency in newer Android rides than in older Android or iOS rides.
This factor could also contribute to the different model evaluation results. 
The achieved \ac{auc} \ac{roc} for uncleaned newer Android data was 0.870, while it was 0.823 for the older Android data.

\begin{figure}[t]
	\centering
	\begin{subfigure}[b]{0.475\textwidth}
		\centering
		\includegraphics[width=\textwidth]{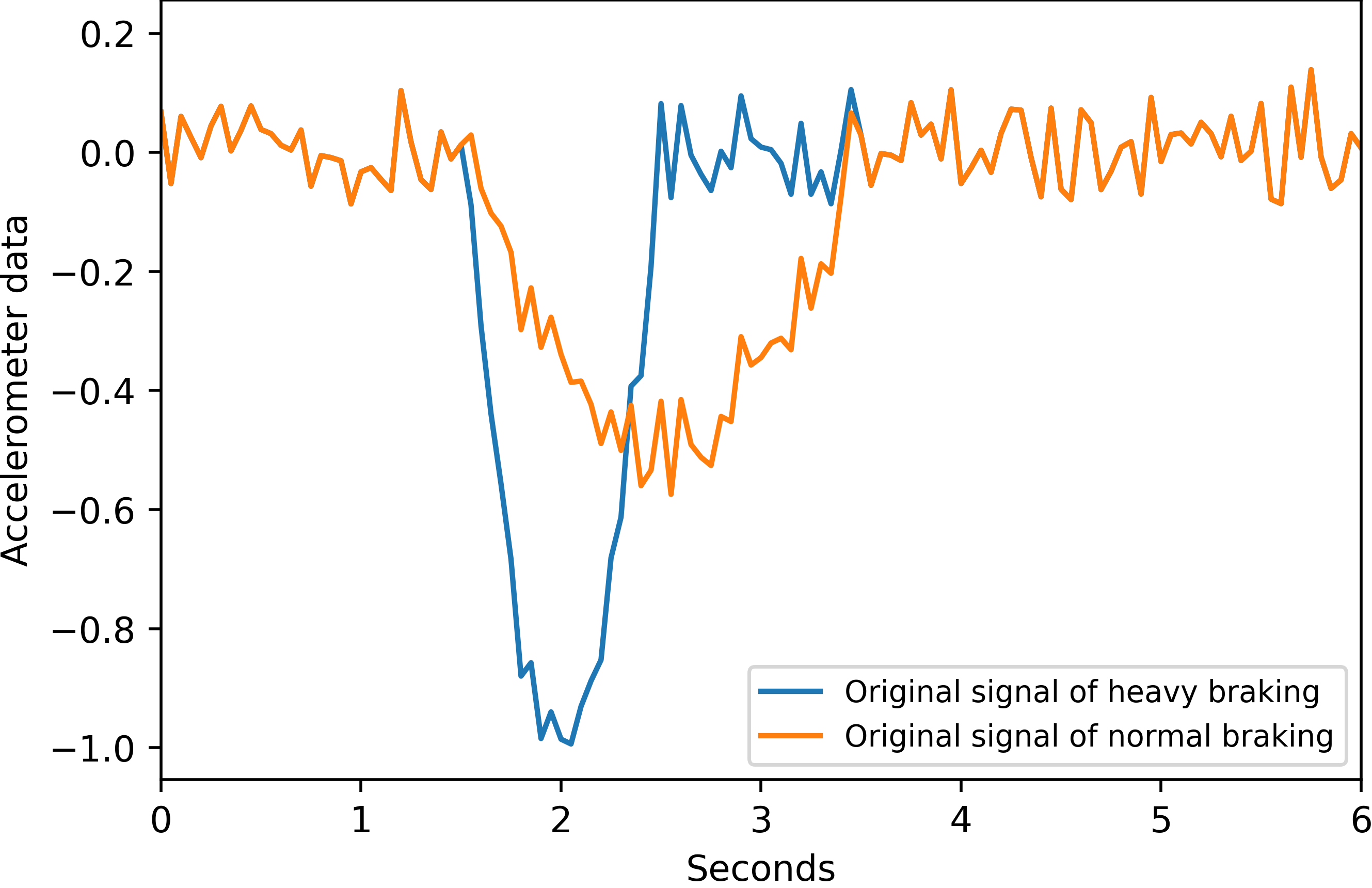}
	\end{subfigure}
	\hfill
	\begin{subfigure}[b]{0.475\textwidth}
		\centering
		\includegraphics[width=\textwidth]{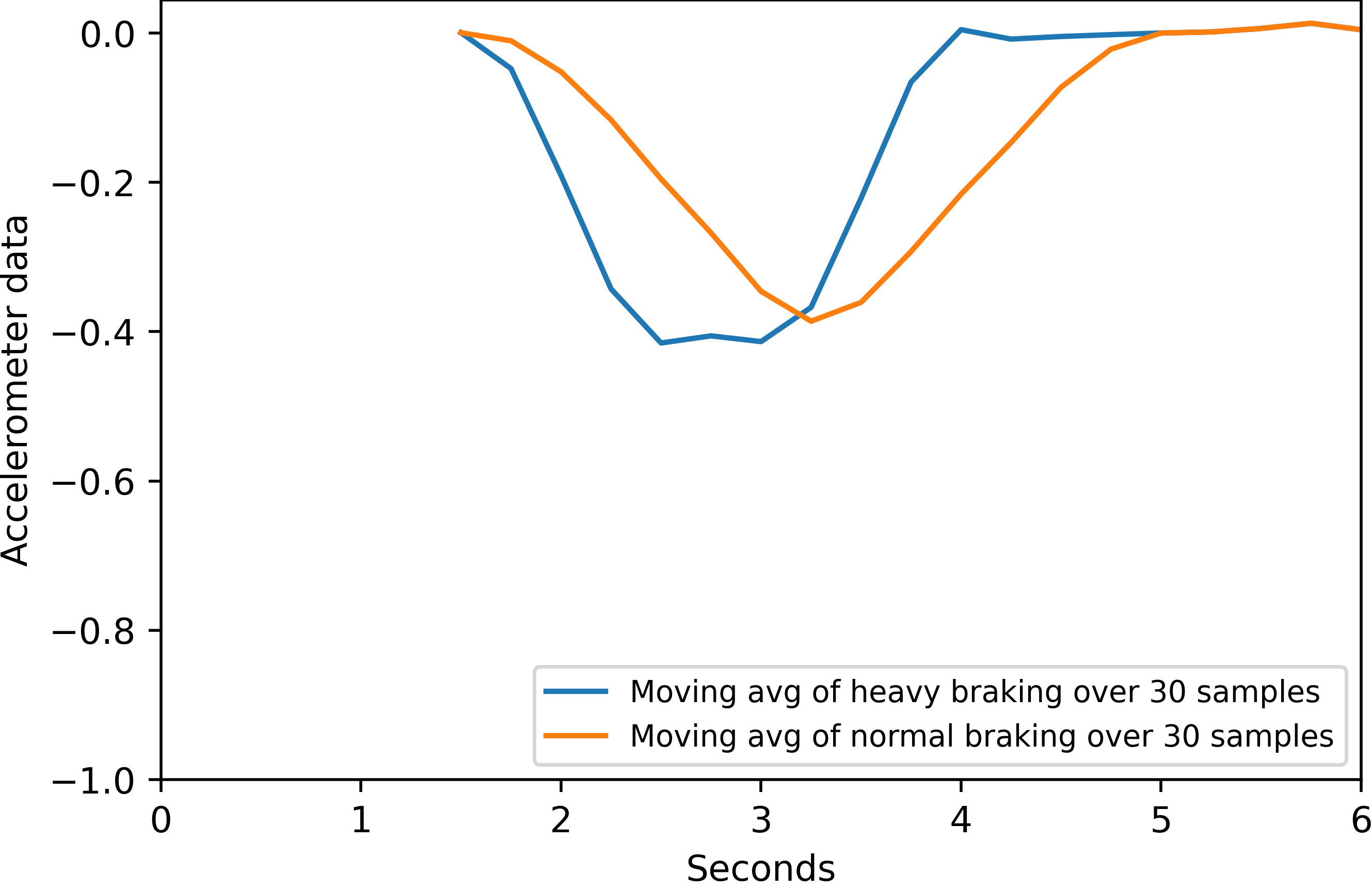}
	\end{subfigure}	
	\caption{Visualization of the sensor data of a simulated heavy braking incident vs.\ a moderate braking event before and after the moving average has been applied.}
	\label{fig:heavy-vs-normal-braking}
\end{figure}

Aside from that, the moving average that is used in the SimRa app to condense the data and comply to users' upload volume constraints~\cite{karakaya2020simra} reduces the amplitude and shifts the exact point in time of incidents and other events.
This has the effect that incidents and non-incidents are hard to distinguish as illustrated in Figure~\ref{fig:heavy-vs-normal-braking}.
Both these factors could hurt the model's ability to classify incidents correctly based on the data.
It is important to acknowledge that this issue becomes less severe when the sampling frequency is higher, as it is the case for the older Android data.

\subsection{Limitation of the Classification Task}

There is an inherent label imbalance present in our data set.
Only relatively few rides contain incidents. 
Moreover, the ratio between timestamps that represent an incident and those that do not is even smaller. 
As a consequence, there is an enormous imbalance between the different label categories.

Furthermore, some incident types such as tailgating or close passes might not be detectable at all in accelerometer and gyroscope sensors~\cite{aldred2018predictors, karakaya2020simra} as cyclists might not change their motion profile despite (or even due to) a dangerous situation.
For detecting these kinds of incidents it may be necessary to include different sensors.
In the SimRa project, this is already happening through an integration of \ac{obs}\footnote{https://www.openbikesensor.org/} data which measures the passing distance of cars.
The current data set, however, only includes very few OBS-supported rides.

In contrast to our classification task, the inherent classification problem of the \ac{har} data set \cite{anguita2013public}, is to classify reoccurring ongoing patterns such as walking or jogging, while an incident might need to be detected by one short non-reoccurring event such as sudden braking.
This further complicates the matter.

\subsection{Data Set Shift}
A major challenge in the context of crowdsourced data are data set shifts. Depending on the device (see also \autoref{subsec:technical_limitations}), but also on other factors, such as the city of origin of the recorded data (see also \autoref{subsec:eval-results}), there could be non-stationarities in the data that violate the assumptions underlying almost all ML models and thus could impact the generalization performance of the trained models. Several types of data set shifts can be distinguished, including \textit{covariate shifts}, meaning shifts in the input data \cite{sugiyama_machine_2012}, \textit{label shifts}, meaning shifts in the distribution of targets \cite{lipton_detecting_2018}, or mixtures thereof. Detecting these shifts \cite{polyzotis_data_2018, rabanser_failing_2018, breck_data_2019, abdar_review_2021, bates_testing_2021} and predicting \cite{schelter_learning_2020} or reducing their impact on the generalization performance is an active field of research \cite{schelter_challenges_2018, biessmann_automated_2021}. Some approaches aim at model specific improvements to alleviate data set shift \cite{sugiyama_machine_2012}. These approaches have a decisive disadvantage, most of these approaches only work for one model class and require access to the inner workings of the ML pipeline, often after the feature extraction step. More promising and easier to build and maintain are model agnostic solutions that focus on the data, rather than the models, to detect and counteract data set shifts \cite{biessmann_automated_2021}. Extending the training data sets to account for all variation and shifts in the data that the ML model should be invariant to, often called \textit{augmentation}, is a popular and effective way of counteracting data set shift, see for instance \cite{cubuk_autoaugment_2019}. In our work we follow this line of thought of a data centric AI approach.
 
\section{Related Work\label{sec:rw}}
In this section, we discuss related work starting with publications related to traffic safety (Section~\ref{subsec:trafficsafety}) before continuing with work regarding \acl{tsc} (Section~\ref{subsec:tsc}).

\subsection{Traffic Safety\label{subsec:trafficsafety}}
Studies on safety in bicycle traffic often rely on crowdsourcing, e.g., using Strava\footnote{https://www.strava.com/} as a data source.
While the Strava data are heavily biased towards recreational trips, some studies rely on them for analyzing various aspects of safety in bicycle traffic, e.g.,~\cite{Hochmair2019,Ferster2021strava}.
Blanc and Figliozzi~\cite{blanc2016modeling, blanc2017safety} study the perceived comfort levels of cyclists based on routes taken.
Wu et al.~\cite{wu2018predicting} predict perceived bicycle safety by combining data from, e.g., OpenStreetMap\footnote{https://www.openstreetmap.com/}, crime statistics, and parking volumes.
Similarly, Yasmin and Eluru~\cite{Yasmin2016} use various open data sources to characterize and provide estimates for bicycle safety in urban areas.
He et al.~\cite{He2018} analyze trajectories of a bike sharing service to detect events of illegally parked vehicles which frequently affect cycling safety.
Figliozzi et al.~\cite{figliozzi2019evaluation} evaluate video recordings to identify safety and delay related problems in bicycle and bus traffic.
Unsurprisingly, bicycles crossing bus lanes can cause slight delays for the buses.

In addition, Kobana et al.~\cite{kobana2014detection} focus on the detection of road damage via smartphone data, and Candefjord et al.~\cite{candefjord2014using} evaluate the development of a crash detection algorithm for cycling accidents.

Most closely related to our work, Aldred and Goodman~\cite{aldred2018predictors} analyze near-miss incidents using road diaries of cyclists and our previous work~\cite{karakaya2020simra} proposed the SimRa platform as well as the heuristic for detecting (near-miss) incidents.
A first extension of our own detection approach~\cite{sanchez2020detecting} developed an \ac{ann} to solve the problem of incident detection on the SimRa data set.
While offering some improvement over the original heuristic, CycleSense clearly outperforms that approach.
Furthermore, Ibrahim et al.~\cite{ibrahim2021cycling} discuss the potential of detecting incidents using image and video data in combination with computer vision techniques.

\subsection{\acl{tsc}\label{subsec:tsc}}
One of the most common tasks for \ac{tsc} are Natural Language Processing, speech recognition, and audio recognition in general.
Another field that deals with this problem is \acf{har} which is concerned with identifying the specific activity of a human based on sensory time series data.
Time windows of a few seconds are classified into activity categories (e.g., walking, sitting, running, lying).
While various feature extraction and pattern recognition methods have been successfully applied in the past in this context~\cite{bulling2014tutorial}, those approaches have constraints such as hand-crafted feature extraction, being able to only learn shallow features~\cite{yang2015deep}, or the requirement for large amounts of well-labeled data for model training~\cite{wang2019deep}.
\acl{dl} techniques and more specifically \acp{cnn} have recently proven to overcome these issues and deliver convincing results in the context of \ac{har}~\cite{wang2019deep,Ronao2015}.
Also, implementations of \acp{rnn} such as \ac{lstm}~\cite{Tao2016,yao2017deepsense} have proven to be successful.
Yao et al.\ \cite{yao2017deepsense} use a combination of \acp{rnn} and \acp{cnn} on top of a \ac{sf} approach after preprocessing their input data using a \ac{dft}.
\section{Conclusion\label{sec:conclusion}}
An increased modal share of bicycles is necessary for solving emission and traffic related urban problems.
A key challenge for this, is the lack of (perceived) safety for cyclists.
Improving the situation requires detailed insights into safety levels of street segments -- the SimRa platform~\cite{karakaya2020simra} has been proposed as a data gathering mechanism for incidents and cycling tracks.
While this is an important step towards data collection, the platform relies on manual annotation of tracks which limits the number of potential users.

In this paper, we have proposed CycleSense -- a model for automatic detection of such incidents.
Using the SimRa data set, we have shown that CycleSense is capable of detecting incidents on the basis of accelerometer and gyroscope time series data in a real-world scenario.
It can correctly distinguish between an incident and a non-incident with a probability of up to 90.5\%.
We have also compared it to the heuristic currently used in the SimRa platform and an existing \ac{fcn} that was specifically developed for SimRa.
Additionally, we have implemented several \ac{dl} models that are frequently used in the context of \ac{tsc}.
We were able to show that our model outperforms all of these approaches.

While this is an important step towards fully automatic incident detection, we believe that -- in the context of the SimRa platform -- CycleSense should be complemented with human annotation in a semi-automated way for the foreseeable future.
Although this does not quite reach our long-term goal of full automation, it should significantly decrease the annotation effort and should also lead to improved labeling quality.
In the future, this could, in turn, be used to further improve CycleSense.
Since April 2022, the SimRa app has been using CycleSense for incident detection.

\bibliography{bibliography}

\end{document}